\documentclass[11pt]{article}

\usepackage[final]{acl}

\usepackage{times}
\usepackage{latexsym}

\usepackage[T1]{fontenc}

\usepackage[utf8]{inputenc}

\usepackage{microtype}

\usepackage{inconsolata}

\usepackage{graphicx}

\usepackage{multirow}        
\usepackage{graphicx}       
\usepackage{booktabs}       
\usepackage{array}          

\usepackage{algorithm}
\usepackage{algorithmic}
\usepackage{amsmath}

\usepackage{amsmath,amssymb,amsfonts}

\usepackage{graphicx}
\usepackage{subcaption}
\usepackage{unicode-math}

\usepackage{xurl}

\newcommand{\std}[1]{{\scriptsize$\pm$#1}}

\title{Mask What Matters: Mitigating Object Hallucinations in Multimodal Large Language Models with Object-Aligned Visual Contrastive Decoding}

\author{
  Boqi Chen\textsuperscript{\rm $\spadesuit$} \quad
  Xudong Liu\textsuperscript{\rm $\diamondsuit$}\thanks{Work done before joining Amazon.} \quad
  Jianing Qiu\textsuperscript{\rm $\clubsuit$}
  \\
  \textsuperscript{\rm $\spadesuit$}ETH Zurich \quad
  \textsuperscript{\rm $\diamondsuit$}Amazon \quad
  \textsuperscript{\rm $\clubsuit$} MBZUAI
  \\
  {\texttt{boqi.chen@ai.ethz.ch}} \quad
  {\texttt{franklxd@amazon.com}} \quad
  {\texttt{jianing.qiu@mbzuai.ac.ae}}
}

\begin{document}
\maketitle
\begin{abstract}
We study object hallucination in Multimodal Large Language Models (MLLMs) and improve visual contrastive decoding (VCD) by constructing an object-aligned auxiliary view. We leverage object-centric attention in self-supervised Vision Transformers. In particular, we remove the most salient visual evidence to construct an auxiliary view that disrupts unsupported tokens and produces a stronger contrast signal. Our method is prompt-agnostic, model-agnostic, and can be seamlessly plugged into the existing VCD pipeline with little computation overhead, \emph{i.e.}, a single cacheable forward pass. Empirically, our method demonstrates consistent gains on two popular object hallucination benchmarks across two MLLMs.
\end{abstract}

\hspace{.5em}\includegraphics[width=1.25em,height=1.25em]{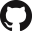}\hspace{.75em}%
\parbox{\dimexpr\linewidth-2\fboxsep-2\fboxrule}{%
  \href{https://github.com/ratschlab/OA-VCD}{%
    \texttt{https://github.com/ratschlab/}\linebreak\texttt{OA-VCD}%
  }%
}

\section{Introduction}
Multimodal Large Language Models (MLLMs) have shown impressive performance acorss various tasks such as image captioning~\citep{li2023blip,qiu2024application} and visual question answering~\citep{lee2024visual,wang2024weakly}, yet they suffer from object hallucination, \emph{i.e.}, mentioning objects not grounded in the image~\citep{li2023pope}. A popular line of work mitigates object hallucination at inference via visual contrastive decoding (VCD). VCD contrasts next-token distributions under the original image and a perturbed auxiliary view to suppress tokens that remain likely without visual support~\cite{leng2024vcd}. Recent works improve VCD by constructing more informative auxiliary views. For instance, VSCoDe~\cite{kim2024vacode} proposes to select the augmentation that maximizes a softmax-distance criterion to strengthen the contrast signal. These perturbations, however, remain heuristic and at image-level, not necessarily aligned with object extents. AGLA~\cite{an2024agla} targets this alignment by preserving prompt-relevant regions while masking distractors using an image–text matching model, and fuse distributions from original and augmented views. Despite good results, it relies on prompt- and model-dependent cross-modal signals, and can risk circularity when biased attention guides the masking intended to correct it.

\begin{figure}
  \centering
  \includegraphics[width=0.99\linewidth]{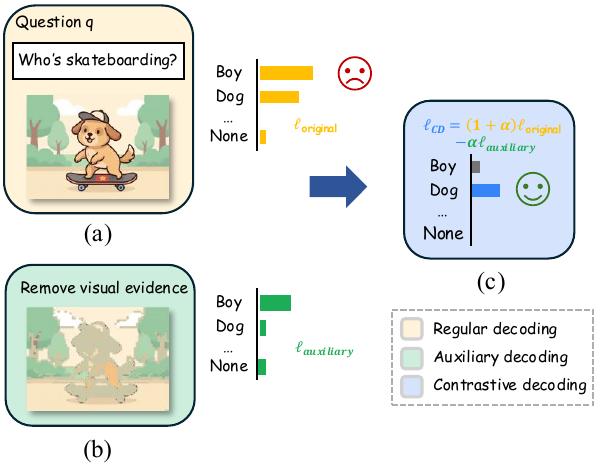}
  \vspace{-1mm}
  \caption{Overview of our method. \textbf{(a)} Regular decoding; \textbf{(b)} decoding using the auxiliary view where visual evidence is removed; \textbf{(c)} contrastive decoding.}
  \label{fig:overview}
\end{figure}

Self-supervised Vision Transformer (ViT) attention maps encode rich cues for semantic segmentation~\cite{dosovitskiy2020image,caron2021dino}. In this paper, we leverage object-centric attention to localize the most salient visual evidence and generate an auxiliary view by masking it out, yielding prompt-agnostic, semantically meaningful counterfactuals that avoid cross-modal dependencies and provide comprehensive object-level perturbations with a single cacheable forward pass. We empirically show that such auxiliary views produce a stronger contrast signal for VCD and leads to better performance on two popular object hallucination benchmarks across two different MLLMs.

\section{Related Work}
\paragraph{Contrastive decoding for reducing object hallucination.} Multimodal generation in MLLMs is prone to object hallucination, \emph{i.e.,} models generate responses that include entities not grounded in the image \citep{rohrbach2018object,li2023pope,wang2023evaluation}. Recent progress therefore targets guided or constrained decoding to improve visual grounding. For instance, VCD contrasts model output distribution using perturbed images to downweight tokens that are driven by language priors~\citep{leng2024vcd}. Following this, several works have proposed more effective ways to generate auxiliary views of the original image. Retrieval VCD retrieves single-concept positive and negative images from a pre-constructed database~\citep{lee2025retrieval}. VSCoDe proposes to select the perturbation from a pool of augmentations (\emph{e.g.}, blur, crop, color) that maximizes a softmax-distance criterion to strengthen the contrast signal~\citep{kim2024vacode}. AGLA combines global context with local discriminative features via an image–prompt matching scheme, highlighting relevant content while mitigating distractors~\citep{an2024agla}. 

Besides VCD, Instruction Contrastive Decoding contrasts output distributions under standard and disturbance instructions (\emph{e.g.}, adding role prefixes) and subtracts the disturbance-induced distribution to detach hallucinated concepts during inference~\citep{wan2024crg}. CODE contrasts between the original self-generated caption and its perturbed variants to identify and suppress hallucinated tokens~\citep{kim2024code}. Activation Steering Decoding applies bidirectional contrastive adjustments to the model’s hidden-state activations during inference—comparing forward and backward pass representations—to steer generated outputs away from hallucinated objects and toward correct ones~\citep{su2025activation}.

\section{Method}
\label{sec:method}
Figure~\ref{fig:overview} provides an overview of our method. In this section, we first review VCD (Section~\ref{vcd}), and then detail how to generate auxiliary views by removing salient visual evidence (Section~\ref{our_method}).

\subsection{Visual Contrastive Decoding}
\label{vcd}
We consider an MLLM with parameters $\theta$. Given a textual query $x$ and a visual input $v$, the model generates a response auto-regressively from
\begin{equation}
\begin{aligned}
y_t \sim\; & p_\theta\big(y_t \mid v,x,y_{<t}\big) \\
\propto\; & \exp\Big(\operatorname{logit}_\theta\big(y_t \mid v,x,y_{<t}\big)\Big).
\end{aligned}
\end{equation}
VCD obtains a second distribution under a \emph{auxiliary} view $v'$ and forms a contrastive distribution that amplifies the differences between the two:
\begin{equation}
\begin{aligned}
p_{\text{vcd}}\big(y \mid v,v',x\big)
&= \mathrm{softmax}\Big( \\
& (1+\alpha)\,\mathrm{logit}_\theta\big(y \mid v,x\big) \\
& -\alpha\,\mathrm{logit}_\theta\big(y \mid v',x\big)\Big).
\end{aligned}
\end{equation}
where $\alpha\ge0$ controls the contrast strength. To avoid penalizing valid outputs from the original distribution and promoting implausible outputs from the
augmented distribution, following \citep{leng2024vcd,an2024agla}, we adopt adaptive plausibility constraints (APC) which selectively consider
tokens with high original probabilities and truncate other
tokens as follows:
\begin{equation}
\begin{aligned}
\mathcal{V}_{\text{head}}(y_{<t})=&\Big\{y_t\in\mathcal{V}:\;
p_\theta\big(y_t \mid v,x,y_{<t}\big) \\
&\ge \beta \max_{w\in\mathcal{V}} p_\theta\big(w \mid v,x,y_{<t}\big)\Big\},
\end{aligned}
\end{equation}
and set $p_{\text{vcd}}\big(y_t \mid v',x,y_{<t}\big){=}0$ if $y_t\notin\mathcal{V}_{\text{head}}(y_{<t})$, with $\beta\in(0,1]$. Combining VCD and APC yields the final decoding rule:
\begin{equation}
\begin{aligned}
y_t \sim \mathrm{softmax}\Big(\,
&(1+\alpha)\,\mathrm{logit}_\theta\big(y_t \mid v,x,y_{<t}\big) \\
&-\alpha\,\mathrm{logit}_\theta\big(y_t \mid v',x,y_{<t}\big)\Big),
\\ &\text{s.t. } y_t\in\mathcal{V}_{\text{head}}(y_{<t}).
\end{aligned}
\end{equation}

\subsection{Generate Auxiliary Views}
\label{our_method}
Given an input image $v\in\mathbb{R}^{H\times W\times 3}$, we extract the attention of the [\texttt{CLS}] token on the heads of the last layer of a self-supervised ViT, \emph{i.e.}, DINO~\citep{caron2021dino}. We then average the attention from multiple heads, reshape it to the patch grid size, and upsample it to $(H,W)$ to obtain a saliency map $\tilde{\mathbf{S}}$, where higher values indicate more prominent visual evidence without task-specific supervision.

To create the an auxiliary view $v'$, we threshold by quantile to remove the regions with high saliency. Formally, let \(\gamma\in(0,1)\) be the area ratio to remove, we define the quantile threshold as:
\begin{equation}
    \lambda_\delta = \mathrm{Quantile}(\tilde{\mathbf{S}},\,1-\gamma),
\end{equation}
and the corresponding binary mask as:
\begin{equation}
\mathbf{M}_\delta \;=\; \mathbf{1}\!\left\{\;\delta\,\big(\tilde{\mathbf{S}}-\lambda_\delta\big) \;>\; 0\;\right\},
\end{equation}
where $\delta=-1$ for removing the most salient region.
Let \(\mathbf{B}\) represent a neutral background (\emph{e.g.}, mean color of neighboring pixels), we obtain an auxiliary view
\begin{equation}
    v' \;=\; \mathbf{M}_\delta \odot \mathbf{B} \;+\; \big(1-\mathbf{M}_\delta\big)\odot v,
\end{equation}
where $\odot$ denoted element-wise multiplication. 

Note that by setting $\delta=1$, the auxiliary view will have the reserve effect, \emph{i.e.}, removing the least salient region (distractors), highlighting the visual evidence as in~\citep{an2024agla}.

\section{Experiments}
\label{exp}

\paragraph{Settings.}
We evaluate on POPE~\cite{li2023pope} and the MME hallucination subset~\cite{yin2024survey} using two different MLLMs: LLaVA-v1.5 (7B) \citep{liu2023llava} and Qwen-VL (7B) \citep{Qwen-VL}. We compare our method against three baselines: regular decoding, VCD with noise-based image perturbation \cite{leng2024vcd} and AGLA \cite{an2024agla}. We threshold at $\gamma=0.8$ and use mean color as neutral background by default (details in Appendix Section~\ref{app-bg-explain}). Ablations on different thresholds $\gamma$ and backgrounds are provided in Section~\ref{exp}. More details on experiment setting can be found in Appendix Section~\ref{app-exp-detail}.

\begin{table}[t]
\centering
\caption{Results (in \%) on the three POPE subsets with LLaVA-v1.5 (7B). Best results are in \textbf{bold}.}
\vspace{-2mm}
\renewcommand\arraystretch{1.05}
\setlength{\tabcolsep}{5.4pt}
\begin{tabular}{l l cc}
\hline\hline
\textbf{Setting} & \textbf{Method} & \textbf{Accuracy $\uparrow$} & \textbf{F1 $\uparrow$} \\
\hline
\multirow{4}{*}{\textit{Random}}     
& Regular   & 84.7 & 83.2 \\
& VCD       & 87.6 & 86.5 \\
& AGLA      & 88.0 & 86.9 \\
& \emph{Ours}     & \textbf{89.5} & \textbf{88.5} \\
\hline
\multirow{4}{*}{\textit{Popular}}     
& Regular   & 80.8 & 79.9 \\
& VCD       & 83.0 & 82.9 \\
& AGLA      & 85.1 & 84.6 \\
& \emph{Ours}     & \textbf{85.7} & \textbf{85.1} \\
\hline
\multirow{4}{*}{\textit{Adversarial}} 
& Regular   & 77.4 & 77.4 \\
& VCD       & 79.4 & 79.9 \\
& AGLA      & 81.2 & 81.3 \\
& \emph{Ours}     & \textbf{81.9} & \textbf{82.0} \\
\hline\hline
\end{tabular}
\label{tab:pope_llava_singlecol}
\vspace{-2mm}
\end{table}

\begin{table}[t]
\centering
\caption{Results (in \%) on the three POPE subsets with Qwen-VL (7B). Best results are in \textbf{bold}.}
\vspace{-2mm}
\renewcommand\arraystretch{1.05}
\setlength{\tabcolsep}{5.4pt}
\begin{tabular}{l l cc}
\hline\hline
\textbf{Setting} & \textbf{Method} & \textbf{Accuracy $\uparrow$} & \textbf{F1 $\uparrow$} \\
\hline
\multirow{4}{*}{\textit{Random}}     
& Regular   & 86.1 & 84.1 \\
& VCD       & 86.7 & 85.0 \\
& AGLA      & 87.4 & 85.7 \\
& \emph{Ours}     & \textbf{88.0} & \textbf{86.5} \\
\hline
\multirow{4}{*}{\textit{Popular}}     
& Regular   & 83.6 & 82.1 \\
& VCD       & 84.0 & 82.5 \\
& AGLA      & 84.8 & 83.8 \\
& \emph{Ours}     & \textbf{85.5} & \textbf{84.3} \\
\hline
\multirow{4}{*}{\textit{Adversarial}} 
& Regular   & 81.1 & 80.0 \\
& VCD       & 81.6 & 80.6 \\
& AGLA      & 82.6 & 81.6 \\
& \emph{Ours}     & \textbf{82.9} & \textbf{82.0} \\
\hline\hline
\end{tabular}
\label{tab:pope_qwen_singlecol}
\vspace{-2mm}
\end{table}

\begin{figure*}[t]
  \centering
  \includegraphics[width=\textwidth]{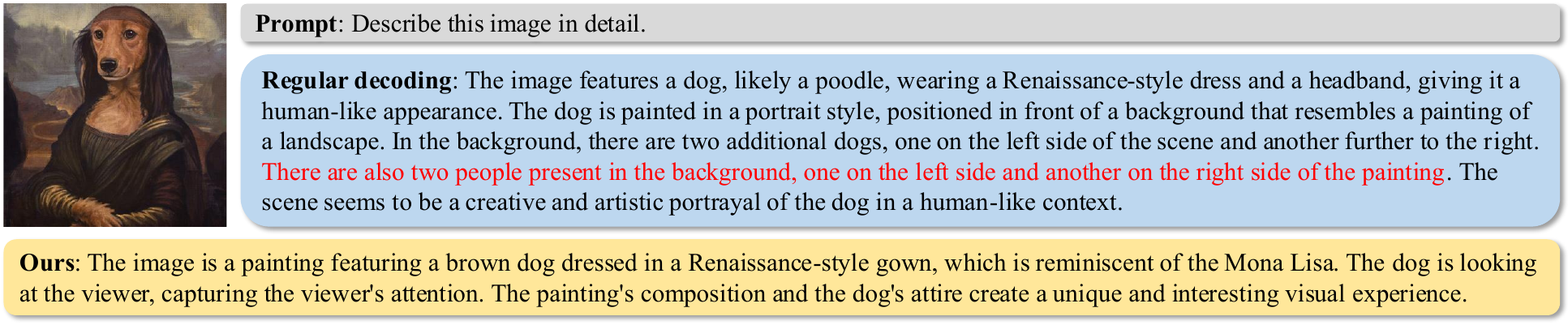}
  \caption{Captions generated by different decoding methods. Hallucinated contents are highlighted in \textcolor{red}{red}.}
  \label{fig:wide}
\end{figure*}

\paragraph{Results.}
Table~\ref{tab:pope_llava_singlecol} and~\ref{tab:pope_qwen_singlecol} reports results on the POPE benchmark using LLaVA-v1.5 (7B) and Qwen-VL (7B), respectively. Across different POPE types and MLLMs, our method consistently improve over baselines. Compared with the strongest baseline, \emph{i.e.}, AGLA, our method achieves higher accuracy and F1 in nearly all settings, with the largest gains on the random subset for LLaVA-v1.5 (7B).

\begin{figure}[h]
\vspace{-3mm}
  \centering
  \includegraphics[width=0.9\linewidth]{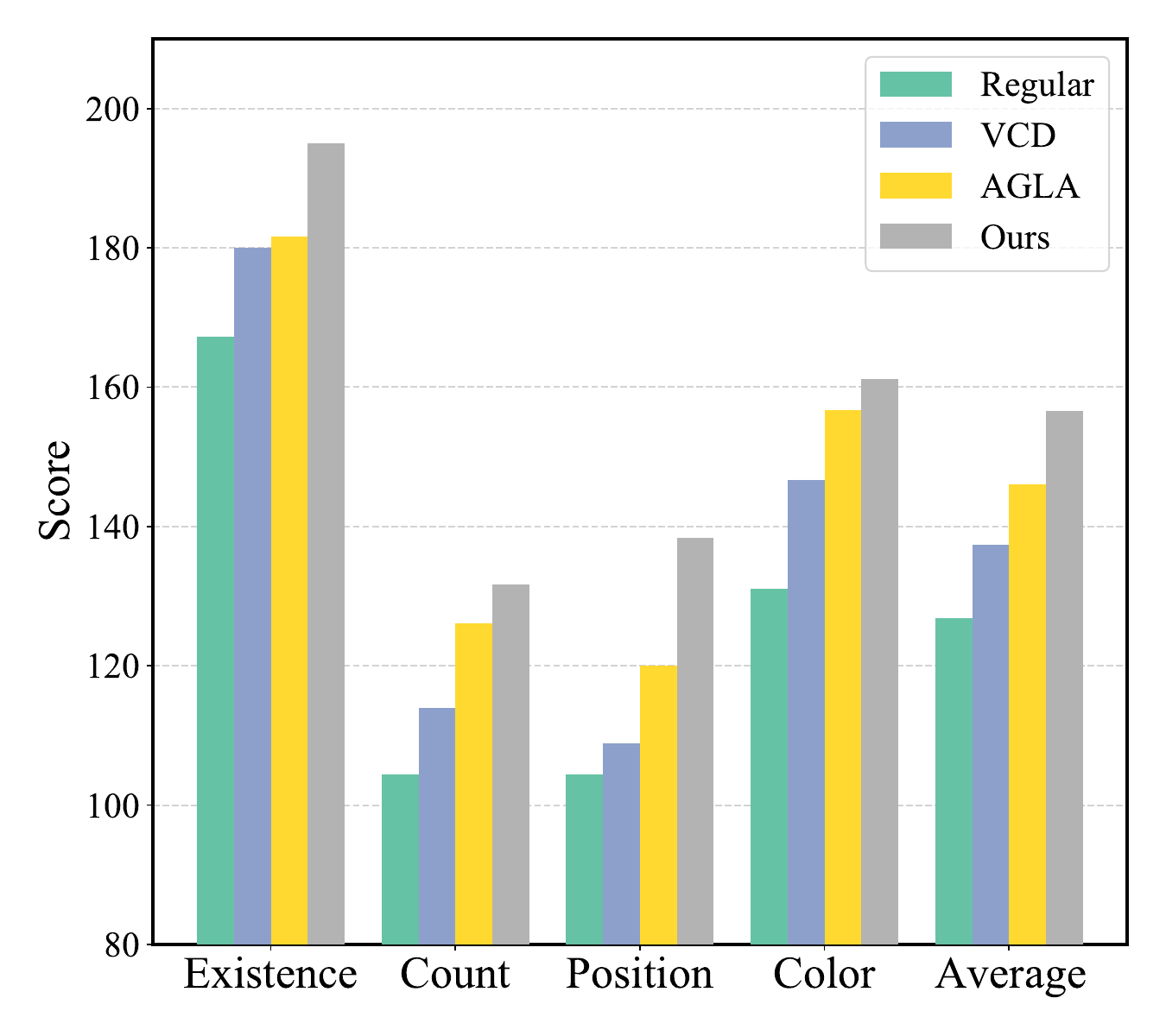}
  \vspace{-2mm}
  \caption{Results averaged across three seeds on the hallucination subset of MME with LLaVA-v1.5 (7B).}
  \label{fig:mme-llava}
\end{figure}

\begin{figure}[h]
\vspace{-3mm}
  \centering
  \includegraphics[width=0.9\linewidth]{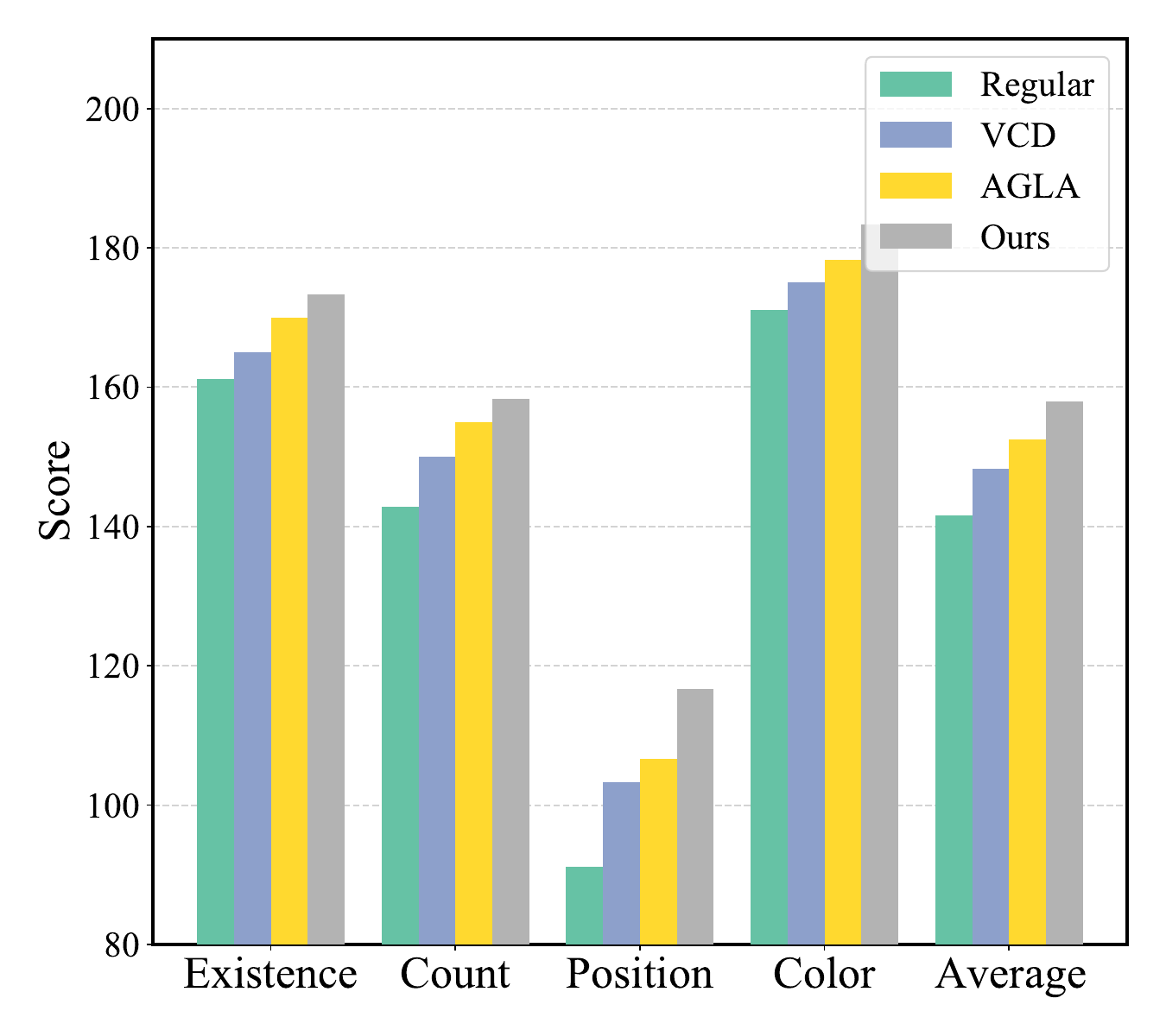}
  \vspace{-2mm}
  \caption{Results averaged across three seeds on the hallucination subset of MME with Qwen-VL (7B).}
  \label{fig:mme-qwen}
\end{figure}

Similar trends hold for the MME benchmark. Figures~\ref{fig:mme-llava} and~\ref{fig:mme-qwen} visualize category-wise scores and the overall average for LLaVA-v1.5 (7B) and Qwen-VL (7B), respectively. Our method achieves the best average score for both models, with the most noticeable improvements on existence and color, the categories that are most susceptible to to spurious object correlations. Overall, these results suggest that masking salient regions from DINO attention yields a stronger, semantically targeted contrast signal while avoiding the prompt dependence of cross-modal masking. The consistent gains across two distinct MLLMs further support the model-agnostic utility of DINO attention for constructing auxiliary views in VCD.

\paragraph{Ablations}
We ablate different thresholds and neutral backgrounds on the POPE MSCOCO subset~\citep{lin2014microsoft}. Overall, we find our method robust to different parameters, with only modest change in performance when varying the thresholds and background methods. 

\paragraph{Case study} 
Figure~\ref{fig:wide} shows a case study on the LLaVA-Bench \citep{liu2023llava}. We can observe that, given the same prompt and image, regular decoding leads to object hallucinations, \emph{e.g.,} "addition dogs" and "two people in the background". We conjecture that these hallucinations stem from the bias and language priors inherent to pretraining. In contrast, our method successfully mitigates these hallucinations without harming the coherence and informativeness of the output caption.

\begin{table}[t]
\centering
\caption{Ablation results (F1, in \%) on the POPE MSCOCO subset with LLaVA-v1.5 (7B) using different backgrounds. Threshold $\gamma=0.8$.}
\vspace{-2mm}
\renewcommand\arraystretch{1.04}
\setlength{\tabcolsep}{6pt}
\begin{tabular}{lccc}
\hline\hline
\textbf{Setting} & Blur & Black & Mean \\
\hline
Random      & 89.4 & 90.7 & 90.5 \\
Popular     & 88.1 & 88.6 & 88.6 \\
Adversarial & 86.0 & 86.4 & 86.2 \\
\hline\hline
\end{tabular}
\label{tab:abl-bg}
\vspace{-2mm}
\end{table}

\begin{table}[t]
\centering
\caption{Ablation results (F1, in \%) on the POPE MSCOCO subset with LLaVA-v1.5 (7B) using different thresholds $\gamma$ with mean background.}
\vspace{-2mm}
\renewcommand\arraystretch{1.04}
\setlength{\tabcolsep}{6pt}
\begin{tabular}{lcccc}
\hline\hline
\textbf{Setting} & 0.2 & 0.4 & 0.6 & 0.8 \\
\hline
Random      & 90.0 & 89.9 & 90.3 & 90.5 \\
Popular     & 88.2 & 88.0 & 88.4 & 88.6 \\
Adversarial & 85.8 & 85.8 & 86.4 & 86.2 \\
\hline\hline
\end{tabular}
\label{tab:abl-thr}
\vspace{-2mm}
\end{table}

\section{Conclusion}
We show that object-aligned auxiliary views, constructed by removing salient visual evidence using a self-supervised ViT’s attention, improve VCD for mitigating object hallucinations in MLLMs. Our method is prompt-agnostic, training-free, and model-agnostic, requiring only a single cacheable forward pass while yielding semantically meaningful, object-level perturbations. Empirically, we demonstrate that such auxiliary views yield a stronger contrastive signal than heuristic augmentations or cross-modal masking, and reduce hallucination on two benchmarks across two MLLMs.

\section*{Limitations}
Our method relies on self-supervised ViT's saliency being well aligned with visual evidence. However, in cluttered scenes, this can lead to under/over-masking. The masking is prompt-agnostic, so in some cases it may suppress regions relevant to the current query, and performance can vary with area ratio and filler choices, though we found a single setting broadly effective. Future work will explore better background filling methods such as diffusion-based image inpainting \citep{corneanu2024latentpaint}.

\section*{Acknowledgement}
This research was primarily supported by the ETH AI Center through an ETH AI Center doctoral fellowship to Boqi Chen.


\bibliography{custom}

\appendix

\section{Appendix}
\label{sec:appendix}


\subsection{Detailed Experiment Settings}
\label{app-exp-detail}
\subsubsection{Benchmarks}

\paragraph{POPE.}
The Polling-based Object Probing Evaluation (POPE) \citep{li2023pope} comprises 27,000 binary queries (\texttt{Yes/No}) targeting object existence across three sources: MSCOCO \citep{lin2014microsoft}, A-OKVQA \citep{schwenk2022okvqa}, and GQA \citep{hudson2019gqa}. For each source, POPE provides three negative-sampling regimes: random, popular, and adversarial. We report Accuracy, Precision, Recall, and F1. For the MSCOCO subset, we adopt the RePOPE annotations \citep{neuhaus2025repope}, which correct erroneous labels and remove ambiguous cases.

\paragraph{MME.}
The MME benchmark \citep{yin2024survey} evaluates broad capabilities of MLLMs, including recognition of object attributes and inter-object relations. In this work, we focus on the hallucination-oriented subset (Existence, Count, Position, and Color), which spans both object- and attribute-level hallucinations. As in POPE, the answers are binary (\texttt{Yes/No}). Following the original protocol, our primary metric is \(\text{Accuracy} + \text{Accuracy}^{+}\), where \emph{Accuracy} is computed per question, and \(\text{Accuracy}^{+}\) is computed per image and requires both associated questions for that image to be answered correctly. The latter is therefore a stricter indicator of comprehensive image-level understanding.

\paragraph{LLaVA-Bench.}
LLaVA-Bench consists of 24 images paired with 60 questions that cover diverse settings, including indoor and outdoor scenes, memes, paintings, and sketches. The benchmark is designed to probe MLLM performance on more challenging problems and out-of-domain scenarios. Following \citet{leng2024vcd}, we present qualitative case studies on this dataset to illustrate the effectiveness of our approach.

\subsubsection{Baselines}
For all baselines, we use the default parameters.

\subsection{Background Methods}
\label{app-bg-explain}
\paragraph{Mean color.}
Let \(\mu\in\mathbb{R}^3\) be the per-channel mean of \(I\) computed over all pixels in normalized space:
\[
\mu_c \;=\; \frac{1}{HW}\sum_{x,y} I_{cxy},\qquad c\in\{1,2,3\}.
\]
The background is the constant field
\[
B_{\text{mean}}(x,y) \;=\; \mu\quad \forall (x,y),
\]
i.e., each masked pixel is replaced by the image’s global mean color (per channel).

\paragraph{Blur.}
Let \(G_{\sigma}\) denote a spatial Gaussian blur (e.g., a \(21\times 21\) kernel). The background is the blurred version of the input,
\[
B_{\text{blur}} \;=\; G_{\sigma}(I),
\]
applied channel-wise. This preserves low-frequency color and illumination while suppressing high-frequency detail inside masked regions.

\paragraph{Black.}
Given normalized RGB inputs,
\[
I_{\text{norm}} \;=\; \frac{I_{\text{rgb}} - \text{mean}}{\text{std}},
\]
define the per-channel constant corresponding to pure black in RGB as
\[
b_c \;=\; -\frac{\text{mean}_c}{\text{std}_c},\qquad c\in\{1,2,3\}.
\]
The background is then
\[
B_{\text{black}}(x,y) \;=\; b \quad \forall (x,y),
\]
which replaces masked pixels with a distribution-consistent black in the model’s normalized space.

\subsection{Dataset License}
POPE, MME, and LLaVA-Bench are intended for research usage.

\begin{itemize}
    \item \textbf{POPE.} It has an MIT License, allowing research usage.
    \item \textbf{MME.} It has a Creative Commons Attribution-ShareAlike 4.0 license, allowing research usage.
    \item \textbf{LLaVA-Bench.} It has a Creative Commons Attribution 4.0 license, allowing research usage.
\end{itemize}

\subsection{Detailed Results on the POPE Benchmark}
The detailed results on the POPE benchmark are shown in Table~\ref{tab:app-1}.
\begin{table*}[t]
\centering
\vspace{-6mm}
\caption{Results (in \%) on the three POPE subsets with LLaVA-v1.5 (7B) and Qwen-VL (7B). Best results are in \textbf{bold}.}
\vspace{-2mm}
\renewcommand\arraystretch{1.02}
\resizebox{0.8\textwidth}{!}{%
\begin{tabular}{cclccc|c}
\hline\hline
\textbf{Model}                         & \textbf{Setting}     & \textbf{Method} & Accuracy $\uparrow$ & Precision & Recall & F1 Score $\uparrow$  \\ \hline\hline
\multirow{12}{*}{LLaVA-v1.5}      
& \multirow{4}{*}{\textit{Random}}     
& Regular           &84.7 &87.3 &79.4 &83.2  \\& 
& VCD              &87.6 &89.1 &84.0 &86.5  \\&  
& AGLA             &88.0 &95.1 &80.2 &86.9  \\&  
& \emph{Ours}             &\textbf{89.5} &92.3 &85.0 &\textbf{88.5}  \\  
\cline{2-7}

& \multirow{4}{*}{\textit{Popular}}     
& Regular           &80.8 &81.1 &78.7 &79.9  \\& 
& VCD              &83.0 &82.4 &83.4 &82.9  \\&
& AGLA             &85.1 &88.1 &81.8 &84.6  \\&  
& \emph{Ours}             &\textbf{85.7} &86.2 &84.1 &\textbf{85.1}  \\  
\cline{2-7}

& \multirow{4}{*}{\textit{Adversarial}}     
& Regular           &77.4 &75.5 &79.4 &77.4  \\& 
& VCD              &79.4 &76.6 &83.7 &79.9  \\&
& AGLA             &81.2 &81.5 &81.7 &81.3  \\&  
& \emph{Ours}             &\textbf{81.9} &79.8 &84.7 &\textbf{82.0}  \\  
\hline\hline

\multirow{12}{*}{Qwen-VL}
& \multirow{4}{*}{\textit{Random}}     
& Regular           &86.1 &91.9 &77.7 &84.1  \\& 
& VCD              &86.7 &91.9 &78.7 &85.0  \\&
& AGLA             &87.4 &93.4 &79.3 &85.7  \\&  
& \emph{Ours}             &\textbf{88.0} &93.4 &80.7 &\textbf{86.5}  \\  
\cline{2-7}

& \multirow{4}{*}{\textit{Popular}}     
& Regular           &83.6 &87.4 &77.4 &82.1  \\&
& VCD              &84.0 &87.8 &78.0 &82.5  \\&
& AGLA             &84.8 &89.6 &78.7 &83.8  \\&  
& \emph{Ours}             &\textbf{85.5} &89.3 &79.8 &\textbf{84.3}  \\  
\cline{2-7}

& \multirow{4}{*}{\textit{Adversarial}}     
& Regular           &81.1 &82.7 &77.5 &80.0  \\& 
& VCD              &81.6 &83.2 &78.1 &80.6  \\&
& AGLA             &82.6 &84.4 &79.0 &81.6 \\&  
& \emph{Ours}             &\textbf{82.9} &84.2 &79.9 &\textbf{82.0}  \\  
\hline\hline
\end{tabular}
}
\label{tab:app-1}
\vspace{-2mm}
\end{table*}

\begin{table*}[h]
\centering
\caption{Results averaged across three seeds on the hallucination subset of MME with LLaVA-v1.5 (7B). Mean and standard deviation are reported. Best results are in \textbf{bold}.}
\label{tab:mme_llava}
\begin{tabular}{lcccc}
\toprule
\textbf{Method} & \textbf{EXISTENCE} & \textbf{COUNT} & \textbf{POSITION} & \textbf{COLOR} \\
\midrule
Regular & 167.22 \std{7.88} & 104.44 \std{1.93} & 104.45  \std{41.41} & 131.11  \std{27.61} \\
VCD & 180.00 \std{0.00} & 113.89 \std{4.41} & 108.89 \std{11.10} & 146.67 \std{22.13} \\
AGLA & 181.67 \std{2.89} & 126.11 \std{0.96} & 120.00 \std{1.67} & 156.66 \std{11.35} \\
\emph{Ours} & \textbf{195.00 \std{5.00}} & \textbf{131.67 \std{8.82}} & \textbf{138.33 \std{4.41}} & \textbf{165.00  \std{3.47}} \\
\bottomrule
\end{tabular}
\end{table*}

\begin{table*}[h]
\centering
\caption{Results averaged across three seeds on the hallucination subset of MME with Qwen-VL (7B). Mean and standard deviation are reported. Best results are in \textbf{bold}.}
\label{tab:mme_qwen}
\begin{tabular}{lcccc}
\toprule
\textbf{Method} & \textbf{EXISTENCE} & \textbf{COUNT} & \textbf{POSITION} & \textbf{COLOR} \\
\midrule
Regular & 161.11  \std{1.92} & 142.78   \std{5.00} & 91.11  \std{9.18} & 171.11  \std{2.55} \\
VCD & 165.00  \std{0.00} & 150.00  \std{2.89} & 103.33  \std{2.89} & 175.00  \std{5.00} \\
AGLA & 170.00  \std{0.00} & 155.00  \std{2.89} & 106.66  \std{2.89} & 178.33  \std{2.89} \\
\emph{Ours} & \textbf{173.33  \std{2.89}} & \textbf{158.33  \std{2.89}} & \textbf{116.66  \std{5.77}} & \textbf{183.33  \std{0.00}} \\
\bottomrule
\end{tabular}
\end{table*}

\subsection{Detailed Results on the MME Benchmark Hallucination Subset}
The detailed results on the hallucination subset of the MME benchmark using LLaVA-v1.5 (7B) and Qwen-VL (7B) are present in Table~\ref{tab:mme_llava} and~\ref{tab:mme_qwen}, respectively. Note that since each type in the hallucination subset only contains 60 questions, resulting 240 question in total, we perform three runs with different randomly seeds and report the average performance for a more robust evaluation.

\subsection{Visualization of Generated Auxiliary Views with Varying Parameters}
\begin{figure*}[t]
  \centering
  \includegraphics[width=0.95\textwidth]{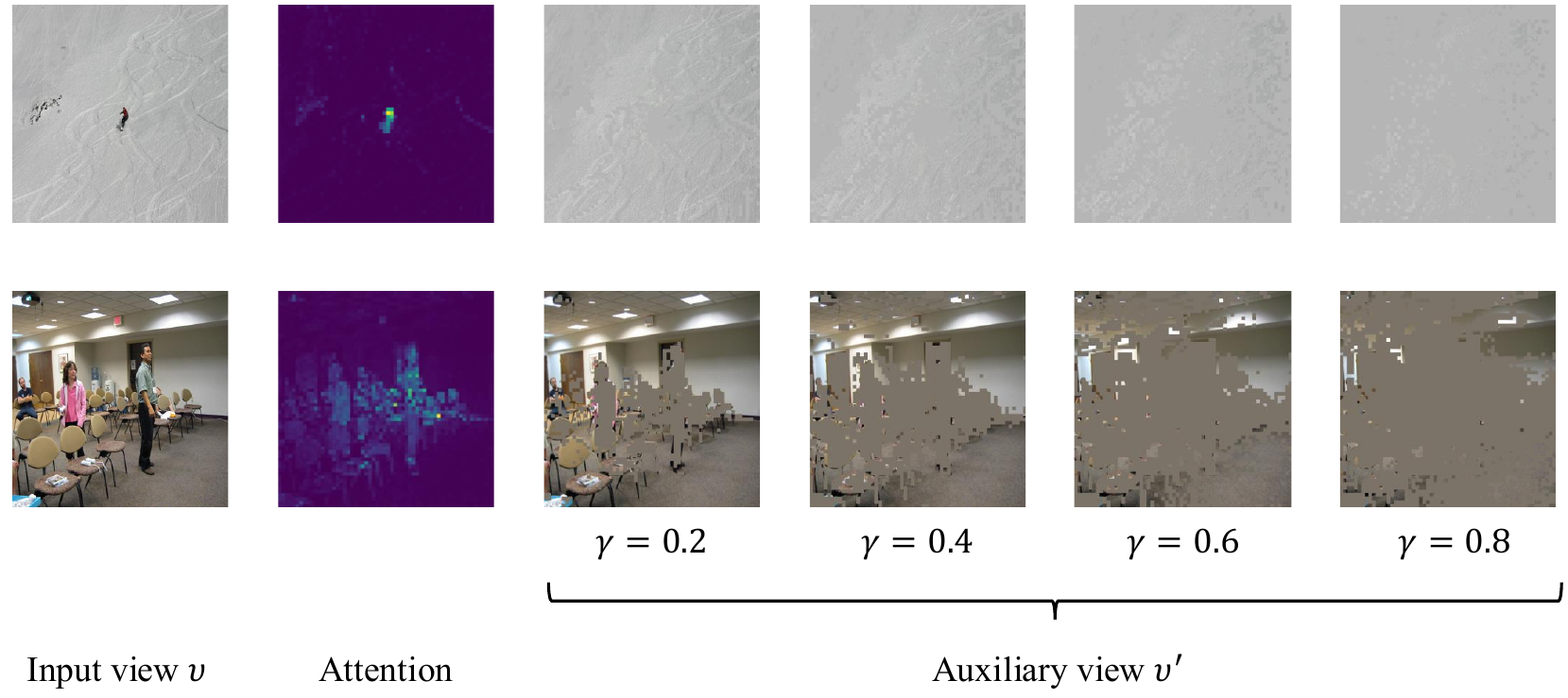} 
  \caption{Visualization of generated auxiliary views with different thresholds. Background are all set to mean color.}
  \label{fig:app-bg}
\end{figure*}

\begin{figure*}[t]
  \centering
  \includegraphics[width=0.95\textwidth]{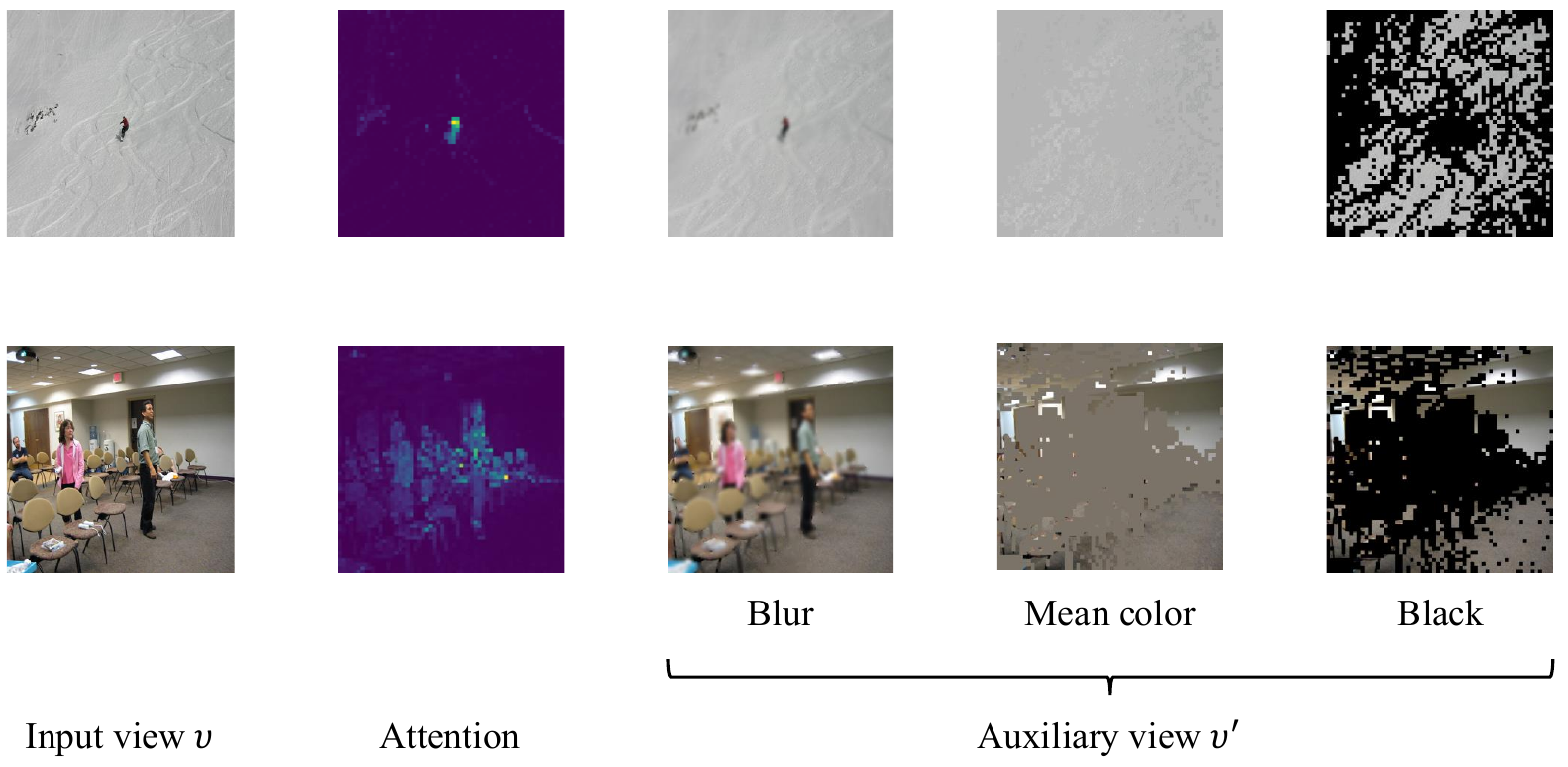} 
  \caption{Visualization of generated auxiliary views with different backgrounds. Thresholds are all set to $\gamma=0.8$.}
  \label{fig:app-thresh}
\end{figure*}

We provide visualizations of the generated auxiliary views by removing visual evidence at different thresholds in Figure~\ref{fig:app-thresh} and with different background inpainting methods in in Figure~\ref{fig:app-bg}.

\end{document}